\def\year{2016}\relax
\documentclass[letterpaper]{article}
\usepackage{aaai16}
\usepackage{times}
\usepackage{helvet}
\usepackage{courier}
\usepackage{graphicx}
\usepackage{caption}
\usepackage{subcaption}
\usepackage{url}
\begin{document}
%
\title{DualNet: Domain-Invariant Network for Visual Question Answering}
\author{Kuniaki Saito, Andrew Shin, Yoshitaka Ushiku, Tatsuya Harada\\
The University of Tokyo\\
7 Chome-3-1 Hongo, Bunkyo\\
Tokyo 113-8654, Japan\\
}
\maketitle
\begin{abstract}
Visual question answering (VQA) task not only bridges the gap between images and language, but also requires that specific contents within the image are understood as indicated by linguistic context of the question, in order to generate the accurate answers. Thus, it is critical to build an efficient embedding of images and texts. We implement DualNet, which fully takes advantage of discriminative power of both image and textual features by separately performing two operations. Building an ensemble of DualNet further boosts the performance. Contrary to common belief, our method proved effective in both real images and abstract scenes, in spite of significantly different properties of respective domain. Our method was able to outperform previous state-of-the-art methods in real images category even without explicitly employing attention mechanism, and also outperformed our own state-of-the-art method in abstract scenes category, which recently won the first place in VQA Challenge 2016.
\end{abstract}

\section{Introduction}
\noindent 
Recent rise of deep learning methods including convolutional neural networks (CNN) and recurrent neural networks (RNN) has escalated a large number of artificial intelligence tasks to an unprecedented stage, where the performance frequently rivals that of humans. Tasks such as object classification, scene classification, and object detection demonstrated the ability to correctly recognize and locate the images both holistically and regionally, whereas tasks such as caption generation or object retrieval demonstrated that deep learning methods can successfully bridge the gap between images and language. Visual question answering (VQA) task further promotes the boundary of deep learning applicability and complicates the problem by necessitating multiple prerequisites, potentially encompassing all of the above-mentioned capabilities; as it needs to understand the question, locate or classify the objects/scenes mentioned in the question, and generate appropriate answers.

In this paper, we introduce DualNet, which attempts to fully exploit the discriminative information provided by the images and textual features, by separately performing addition and multiplication of input features to form a common embedding space. As we shall see in Experiment Section, it shows clear advantage over performing only one operation, and outperforms many recent state-of-the-art methods, without using any attention mechanism. Furthermore, it turns out that building an ensemble of DualNets with varying dimensions leads to even more superior performances, despite feeding identical set of input features to all DualNet units.

Another advantage of our DualNet is that it is applicable to both real images and abstract scenes categories. So far, it has widely been considered that successful methods for real images cannot be directly ported to abstract scenes domain, as they have fundamentally different characteristics. In fact, applying the basic setting of fc7 features for images and long short-term memory (LSTM) with one hidden layer for questions, which results in 58.16 for real images, yields only about 55 in abstract scenes domain. Indeed, most of the previous papers on VQA have tackled only one domain, presumably due to such reason. Our DualNet, however, results in superior performances in both domains, demonstrating that it is applicable to a wider domain, provided that features are plausible.

It is also noteworthy that we do not employ any attention mechanism, which has become one of the most common approaches in VQA. While useful, building attention mechanism necessitates a separate stage of training to map language to specific regions of image, and complicates the procedure. Instead, our method demonstrates that basic set of features can provide rich amount of information without building attention mechanism, provided a network is designed in a way that fully exploits the features' discriminative capacity.

This paper is hereafter structured as follows; In Related Works, we review the recent innovations and trends in VQA, and briefly discuss how our method diverts from them. In Method, we describe both the motivation behind and the implementation details of our DualNet architecture. In Experiment, we apply our proposed model to actual VQA dataset and discuss the results with examples and comparisons to other methods. Finally, we conclude the paper and discuss future work in Conclusion.

\section{Related Work}
Visual question answering (VQA) task itself has only recently been introduced with the advent of dataset provided by \cite{Antol_2015_ICCV}, consisting of 0.25M images, 0.76M questions, and 10M answers. They also report baseline results from methods with multi-layer perceptron and LSTM \cite{LSTM}.

\textbf{VQA: Real Images} Real images category is currently by far the more popular and competitive task in VQA. \cite{malinowski2016ask} introduced Ask Your Neurons. Unlike the baseline provided by \cite{Antol_2015_ICCV}, in which image features and question features are embedded to common space at the last stage prior to classification, they built a system where image features are shared at each LSTM unit for processing question features. They also performed comparison of different operations for fusing input features, and concluded that summation performs better than multiplication. In our work, however, both summation and multiplication are performed, which demonstrates significant improvements.

Many recent papers reporting competitive results have relied heavily on various types of attention mechanism. \cite{yang2015stacked} introduced stacked attention networks (SANs), which relies on semantic representation of each question to search for relavant regions in the image. More specifically, they built multiple-layer attention mechanism, which locates the relevant region multiple times so that more accurate region of interest can be retrieved.

In a similar manner, \cite{Shih} attempts to locate relevant regions in the image. They map the textual queries to features from different regions by embedding them to a common space and comparing their relevance via inner product.

\cite{xiong2016dynamic} proposed a number of improvements to dynamic memory network (DMN). Their proposed DMN+ model introduced a novel input module based on a two-level encoder with sentence reader and input fusion layer, and implemented memory based on gated recurrent units (GRU). 

\cite{ilievski2016focused} proposed focused dynamic attention (FDA) model, which exploits an object detector to determine regions of interest. LSTM is used to embed the region features and global features into common space. 

\cite{Ask} proposed spatial memory network in which neuron activations of different spatial regions are stored in memory, and regions with high relevance are chosen depending on the question. The latter step was made possible by their novel spatial attention architecture designed to align words with patches.

Unlike most of the works mentioned above, our work does not employ any attention mechanism, yet demonstrates superior performance by fully exploiting features provided to the network.

\textbf{VQA: Abstract Scenes} Relatively few results have been reported on abstract scene categories compared to real images. 

\cite{zhang} converted the questions to a tuple containing essential clues to the visual concept of the images. Each tuple (P, R, S) consists of a primary object (P), secondary object (S), and their relation (R). Mutual information was employed to determine which object corresponds to primary object and secondary object. They also augmented the dataset using crowd-sourcing in order to balance the biases in the dataset. Their visual features included histogram-like vectors for primary and secondary objects, as well as absolute and relative locations of the objects modeled by GMMs. We show that this model's performance is enhanced by addition of deep features, both holistically and regionally, and applying our DualNet further improves the performance.

\section{Method}
In this section, we describe the details of our proposed network architecture ``DualNet", which we demonstrate to work well both on real images and abstract image. Furthermore, we demonstrate that it performs well on various combinations of image features when combined with sentence features from encoders such as LSTM.

\subsubsection{Motivation}
In the VQA task, it is necessary to determine how to combine visual features with sentence features because a network cannot answer correctly unless they have enough knowledge about what the questions are asking and which features are necessary to answer them correctly. Figure 1 shows an example of fusing features which employs element-wise multiplication \cite{Antol_2015_ICCV}. There are other options to fuse the features such as element-wise summation. Some of the previous works have examined and compared the behaviors of network depending on the fusing mechanisms \cite{malinowski2016ask}. According to them, the performance of network varies depending on the way the image features and sentence features are fused. This indicates that, even with non-linearity of network, the information can vary according to the fusing methods. Most architectures only used one method to fuse the features; for example, summation or multiplication only.

However, the features combined with different fusing methods should contain different information. For example, some information should be preserved (or lost) only by summation, whereas some are preserved only by multiplication. For this reason, we propose to integrate two kinds of operations, namely element-wise summation and element-wise multiplication. Moreover, we propose to use different kinds of image features. The motivation is to fully take advantage of different information present in different kinds of features. For example, holistic features used in abstract scenes \cite{zhang} display completely different characteristics from CNN features. Likewise, for CNN features, different network structures also result in different characteristics of the extracted features. Thus, our DualNet benefits further by exploiting a combination of features from different networks and different methods.

\begin{figure*}[htbp]
 \begin{minipage}{0.25\hsize}
  \begin{center}
   \includegraphics[width=\hsize]{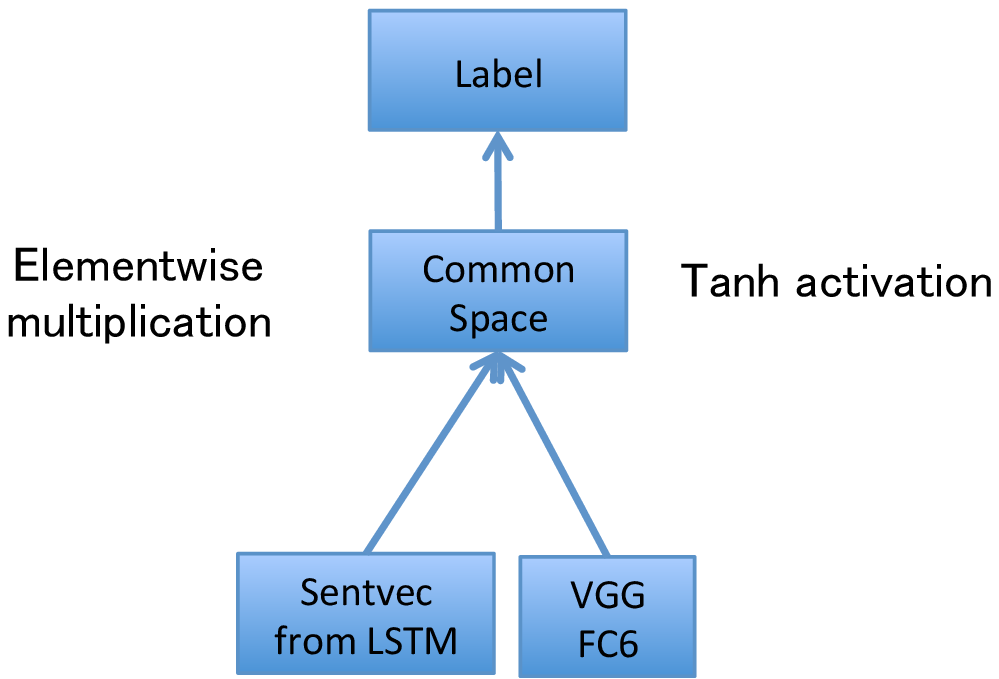}
  \end{center}
    \caption{Basic network architecture for VQA}
  \label{fig:one}
 \end{minipage}
 \begin{minipage}{0.38\hsize}
  \begin{center}
   \includegraphics[width=\hsize]{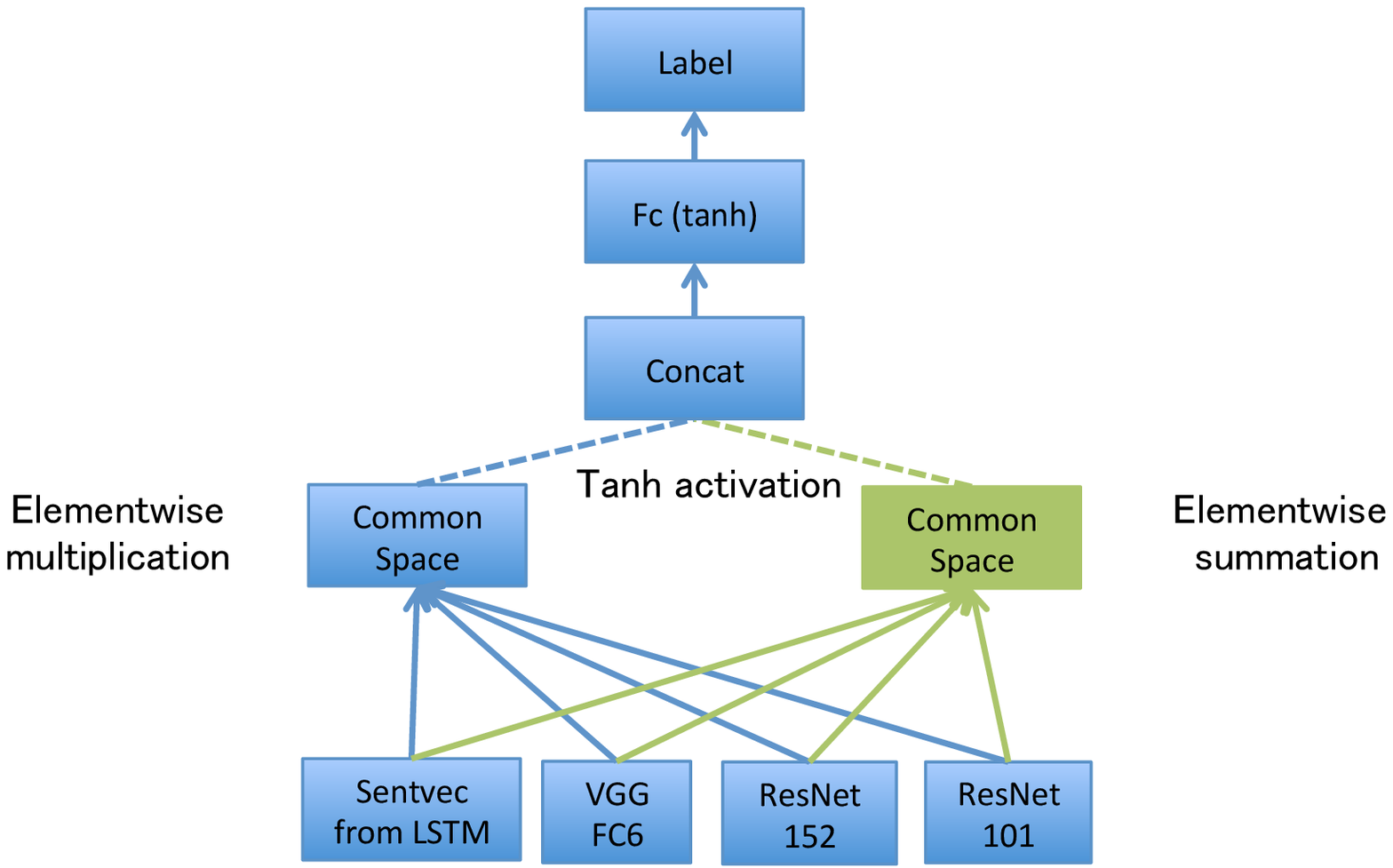}
  \end{center}
  \caption{DualNet for real images}
  \label{fig:two}
 \end{minipage}
 \begin{minipage}{0.38\hsize}
  \begin{center}
   \includegraphics[width=\hsize]{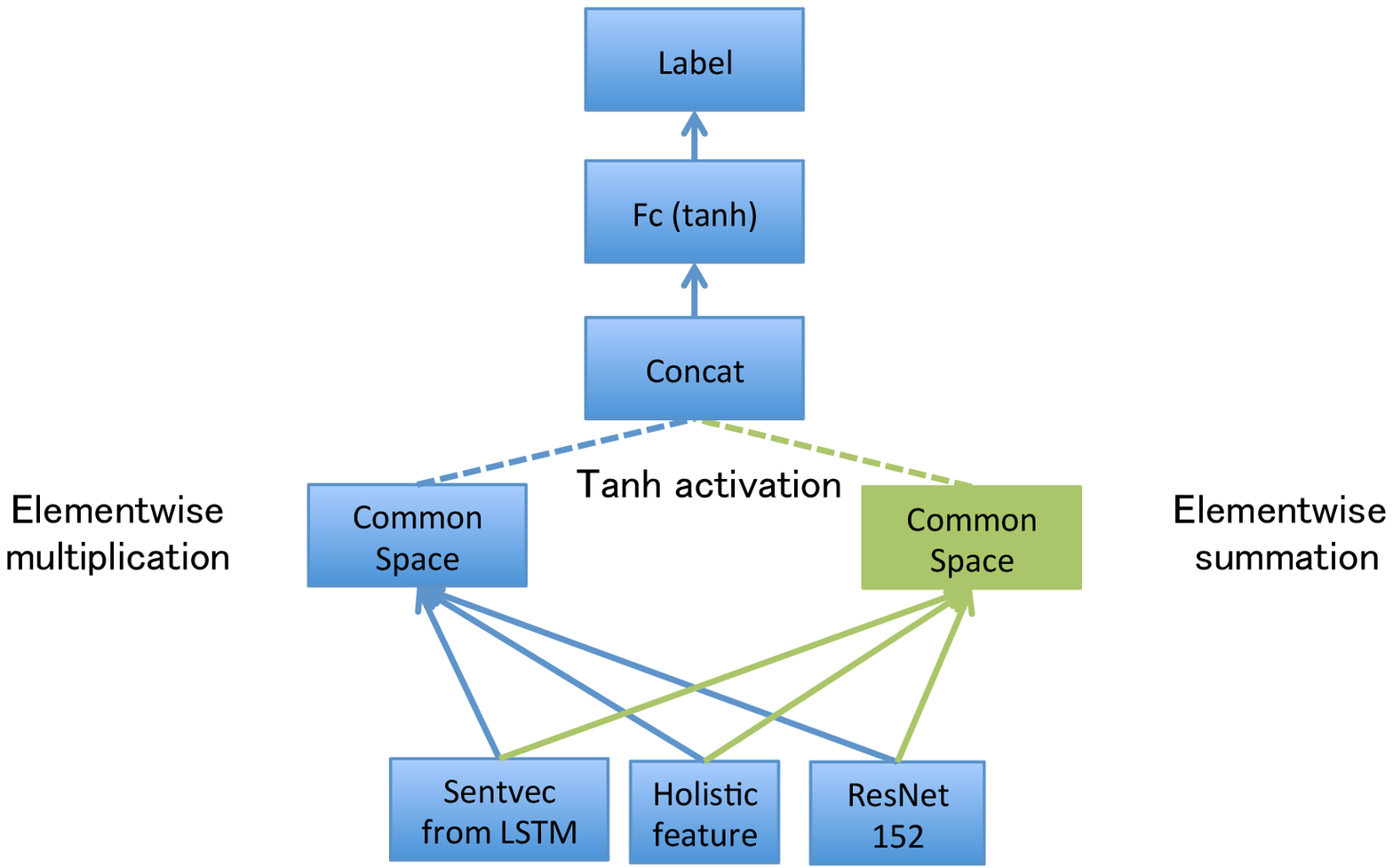}
  \end{center}
  \caption{DualNet for abstract scenes}
  \label{fig:three}
 \end{minipage}
\end{figure*}

\subsubsection{Implementation}
We now go through the theoretical background of our network. We skip notations for bias parameters in the following equations for clarity.
\begin{eqnarray}
Q = LSTM(x_1,x_2,x_3,...x_t)
\end{eqnarray}
First, we input one-hot vector of words sequentially and obtain question vector from the last hidden layer of LSTM.
\begin{eqnarray}
{I_M}_1^{'} = tanh({W_M}_1I_{1})\\
{I_M}_2^{'} = tanh({W_M}_2I_{2})\\
{Q_M}^{'} = tanh({W_M}_qQ)\\
F_M = {I_M}_1^{'}\circ{I_M}_2^{'}\circ{Q_M}^{'}
\end{eqnarray}
Eq. (2) to (5) correspond to the fusing of image features and text features by multiplication. ${\circ}$ refers to the element-wise multiplication.
\begin{eqnarray}
{I_S}_1^{'} = tanh({W_S}_1I_{1})\\
{I_S}_2^{'} = tanh({W_S}_2I_{2})\\
{Q_S}^{'} = tanh({W_S}_qQ)\\
F_S = {I_S}_1^{'}+{I_S}_2^{'}+{Q_S}^{'}
\end{eqnarray}
Eq. (6) to (9) correspond to the fusing of the features by summation. Our proposed network does not share the weight between multiplication and summation because we expect each operation to extract different kinds of information.
\begin{eqnarray}
F = Concat(F_M,F_S)\\
Output = {W_f}_2tanh({W_f}_1F)
\end{eqnarray}
We concatenate the features from element-wise multiplication and element-wise summation. In this example, we have shown the case where we use two kinds of image features. We can change the number of image features depending on the needs, but the overall workflow will remain the same regardless of the number of features.

The proposed model architecture for real image is described in Fig.2. It uses L2-normalized features from the first fully-connected layer (fc6) of VGG-19\cite{vgg} extracted using Caffe \cite{caffe} trained on ImageNet\cite{ImageNet}, and the uppermost fully-connected layers from ResNet-152 and ResNet-101 \cite{resnet} for image features, in order to construct DualNet. The proposed model architecture for abstract image is described in Fig.3. It uses L2-normalized holistic feature, and fully-connected layer of ResNet-152 for image features.

\begin{table*}
\begin{center}
 \caption{Performances of each method on test-dev split of real images category}
  \begin{tabular}{cccccccccccccccc}
     \hline
   &\multicolumn{4}{c}{Open-Ended}&&\multicolumn{4}{c}{Multiple-Choice}\\\cline{2-5}\cline{7-10}
       &All &Y/N&Num&Others&&All&Y/N&Num&Others\\
    \hline
    DPPnet \cite{noh2015image} & 57.22&80.7&37.2&41.7&&62.50&80.8&38.9&52.2\\
      deeper LSTM Q+norm \cite{Lu2015}& 57.75&80.5&36.8&43.1&&62.70&80.5&38.2&53.0\\
        SAN \cite{yang2015stacked}& 58.70 &79.3 &36.6 &46.1&&-&-&-&-\\
            FDA\cite{ilievski2016focused}&59.24&81.1&36.2&45.8&&64.01&81.5&39.0&54.7\\
         DMN+\cite{xiong2016dynamic}&60.30&80.5&36.8&48.3&&-&-&-&-\\\hline
      Sum only& 56.81&78.4&35.2&43.3&&-&-&-&-\\
      Mul only  &59.15&80.6&37.0&45.8&&-&-&-&-\\\hline
      DualNet &60.47&81.0&37.1&48.2&&65.80&80.8&39.8&58.9\\
      DualNet (ensembled)& {\bf 61.47}&{\bf 82.0}&{\bf 37.9}&{\bf 49.2}&&{\bf 66.66}&{\bf82.1}&{\bf39.8}&{\bf59.5}\\\hline
       \end{tabular}
             \label{table:result_dev}
            \end{center}
  \end{table*}
 
\begin{table*}
\begin{center}
\caption{Performances of each method on test-std split of real images category}
 \begin{tabular}{cccccccccccccccc}
    \hline
  &\multicolumn{4}{c}{Open-Ended}&&\multicolumn{4}{c}{Multiple-Choice}\\\cline{2-5}\cline{7-10}
      &All &Y/N&Num&Others&&All&Y/N&Num&Others\\
   \hline
   DPPnet \cite{noh2015image} & 57.36&80.28&36.92&42.24&&62.69&80.35&38.79&52.79\\
   D-NMN \cite{andreas2016learning}  & 58.0 & -&- &-&&-&-&-&-\\
   deeper LSTM Q+norm \cite{Lu2015}& 58.16&80.56&36.53&43.73&&63.09&80.59&37.70&53.64\\
   AYN \cite{malinowski2016ask}&58.43 &78.24&36.27&46.32&&-&-&-&-\\
       SAN \cite{yang2015stacked}& 58.90 & -&- &-&&-&-&-&-\\
    FDA \cite{ilievski2016focused} & 59.54 & 81.34&35.67 &46.10&&64.18&81.25&38.3&55.20\\
   DMN+ \cite{xiong2016dynamic}& 60.36 & 80.43&36.82 &48.33&&-&-&-&-\\\hline
         DualNet (ensembled) & {\bf 61.72} & {\bf 81.92}&{\bf 37.84} &{\bf 49.66}&&{\bf 66.72}&{\bf 81.95}&{\bf 39.72}&{\bf 59.55}\\\hline
      \end{tabular}
            \label{table:result_std}
           \end{center}
 \end{table*}

\section{Experiment}
In this section, we describe and discuss the results from the experiments using our DualNet architecture on VQA dataset. 

\subsection{Real Images}
The dataset consists of 82,783 training images, 40,504 validation images and 81,434 test images. 3 questions are attached to each image. We can evaluate the model by using a subset of test split called test-dev split on the VQA evaluation server. In this experiment, we used both train and validation splits for training, and tested on both test-dev and test splits. Since the number of test submissions for the complete test split is limited, the evaluation on the complete test split was restricted to selected key methods.

LSTM in our model consists of 2 layers with 512 hidden units. We used 2,000 most frequent answers as labels, and relied on rms-prop to optimize our model. The batch size was 300 and learning rate was set to 0.0004. We optimized the hyper-parameters based on the evaluations on test-dev split.

The model performance slightly changed according to the dimension of common space. We show the result of 1024 dimension as single DualNet's result.
For the ensemble of DualNets, we set the common space dimensions differently for each unit. We changed common space dimension from 500 to 3000 for each DualNet unit in our ensemble, which consists of 19 DualNet units. We tuned the weight for each unit in the ensemble based on their result on test-dev split.

\subsection{Abstract Scenes}
Abstract scenes category contains 20,000 images for training, 10,000 images for validation, and 20,000 images for test images, where each image is accompanied by 3 questions. Unlike real images, there is no test-dev split. 

So far, it has widely been believed that abstract scenes possess fundamentally different properties from those of real images, and thus successful methods for real images cannot be directly ported to abstract scenes, necessitating a significantly different approach.

\subsubsection{Baseline 1}
As our first baseline, we follow Zhang et al. \cite{zhang} whose method was described in Related Works. 

\subsubsection{Baseline 2}
We now describe the second baseline also implemented by ourselves, which recently won the first place in VQA Challenge 2016, and is currently the state-of-the-art method in the abstract scenes category.

On top of the features described in \cite{zhang}, we added features from the uppermost fully-connected layer from ResNet with 152 layers, and fc7 layer of VGG with 19 layers for holistic features. We alternated between two different setups for regional features as following:

1) Avg. Softmax of Top Regions: we first extract 10 regions from each image using Deep Proposal \cite{deepproposal}, which proposes regions based on objectness measure and applies non-maximum suppression to filter out overlaps. We then extract softmax probabilities for each region, which correspond to 201 classes used in ILSVRC object detection task. We used Fast-RCNN \cite{fastrcnn} and VGG-16 trained for the task. Finally, we average the softmax probabilities of all 10 regions to obtain one 201-dimensional vector.

2) VLAD Coding of CNN with Coordinates: The general procedure is similar to \cite{vlad} except we do not employ spatial pyramid. We run selective search for each image, which returns approximately 1,000 region proposals for each image. Using Fast-RCNN, we extract fc7 features from all regions. Dimensionality of fc7 features is reduced to 256 using PCA. We then concatenate 8-dimensional coordinate vector (x\_min, y\_min, x\_max, y\_max, x\_center, y\_center, width, height) as in \cite{retrieval} so that each region is 264-dimenisonal. Finally, we apply VLAD coding to all regions of an image with one cluster to obtain the final one 264-dimensional vector for each image. 

It turns out that 1) performs better on yes/no and number questions, while 2) performs better on others category. We thus alternated between the two methods depending on the type of question, which was predicted by key phrase extraction; e.g., `how many' indicating number category, etc. We had batch size of 400, and 500 possible answers, and set number of word embeddings for questions as 1,000. LSTM with one hidden layer of 256 hidden units was employed. Training was performed for 100 epochs.

	 \begin{table*}
\begin{center}
 \caption{Performances of each method on test data of abstract scenes category}
  \begin{tabular}{cccccccccccccccc}
     \hline
   &\multicolumn{7}{c}{Open-Ended}&&\multicolumn{7}{c}{Multiple-Choice}\\\cline{2-8}\cline{10-16}
       &All &&Y/N&&Num&&Others&&All&&Y/N&&Num&&Others\\
    \hline
      Baseline1 \cite{zhang}& 65.02&&77.5&&52.5&&56.4&&69.21&&77.5&&52.9&&66.7\\
       Baseline 2 & 67.39&&79.6&&57.1&&58.2&&71.18&&79.6&&56.2&&67.9\\
       MRN \cite{kim2016multimodal}&62.56&&79.1&&51.6&&48.9&&67.99&&79.1&&52.6&&62.0\\\hline
      DualNet &68.87&&80.0&&57.9&&61.1&&73.29&&80.0&&58.5&&71.8\\
      DualNet (ensembled)&{\bf69.73}&&{\bf 80.7}&&{\bf 58.8}&&{\bf 62.1}&& {\bf74.02}&&{\bf 80.8}&&{\bf59.2}&&{\bf 72.4}\\\hline
       \end{tabular}
             \label{table:result_dev}
            \end{center}
  \end{table*}
  
\begin{figure*}[t!]
    \centering
    \begin{subfigure}[t]{0.3\textwidth}
        \centering
        \includegraphics[height=1.2in]{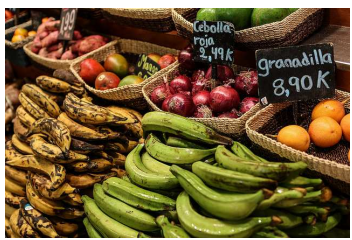}
        \caption{\textbf{Q}: What fruit is yellow and brown? \\ \textbf{A}: banana}
    \end{subfigure}%
    ~ 
    \begin{subfigure}[t]{0.3\textwidth}
        \centering
        \includegraphics[height=1.2in]{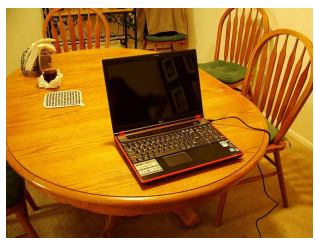}
        \caption{\textbf{Q}: Is this a laptop? \ \textbf{A}: yes}
    \end{subfigure}
        ~ 
    \begin{subfigure}[t]{0.3\textwidth}
        \centering
        \includegraphics[height=1.2in]{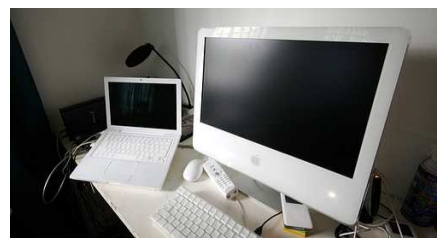}
        \caption{\textbf{Q}: How many screens are there? \ \textbf{A}: 2}
            \end{subfigure}
    \caption{Examples of question and generated answers in real images}
\end{figure*}

\begin{figure*}[t!]
    \centering
    \begin{subfigure}[t]{0.3\textwidth}
        \centering
        \includegraphics[height=1.2in]{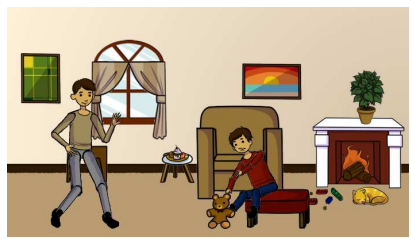}
        \caption{\textbf{Q}: What is the boy playing with? \\ \textbf{A}: teddy bear}
    \end{subfigure}%
    ~ 
    \begin{subfigure}[t]{0.3\textwidth}
        \centering
        \includegraphics[height=1.2in]{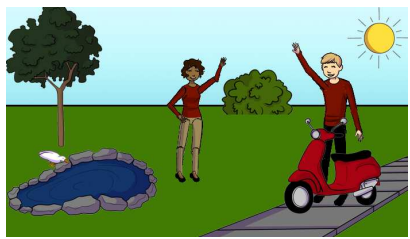}
        \caption{\textbf{Q}: Are there any animals swimming in the pond? \ \textbf{A}: No}
    \end{subfigure}
        ~ 
    \begin{subfigure}[t]{0.3\textwidth}
        \centering
        \includegraphics[height=1.2in]{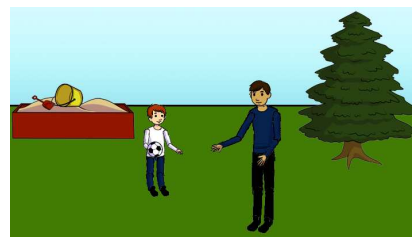}
        \caption{\textbf{Q}: How many trees? \ \textbf{A}: 1}
    \end{subfigure}
    \caption{Examples of question and generated answers in abstract scenes}
\end{figure*}

\subsection{Results \& Analysis}



Table 1 shows the results of each method on test-dev split, and Table 2 reports results on test-std split.
As shown in the tables, we outperformed the previous state-of-the-art methods on real image category published prior to the 2016 VQA Challenge.
We can clearly see the effectiveness of our network structure through comparison to the summation-only network and multiplication-only network. The multiplication network
obtained 59.15 and summation network obtained only 56.81. The  performance of summation network is much poorer than multiplication network. However, when combining two
paths, we were able to improve the performance significantly. This indicates that the two paths extract different kinds of information from more than three kinds of features, reminiscent of the way "And" and "Or" gates behave in electronic circuits. 

Comparing with the methods such as DMN \cite{xiong2016dynamic} , SAN \cite{yang2015stacked} and FDA \cite{ilievski2016focused}, which used the attention mechanism, our model still achieves better performance. This indicates that we can construct an efficient model without explicitly including spatial information from local features. It also suggests that the image features from VGG and ResNet must contain spatial information to a useful extent, since our model demonstrates high performance on the questions that require the model to have knowledge about particular image regions in order to answer correctly. Figure 4 shows examples of questions and generated answers in real images along with the images.

As for abstract image, our DualNet method significantly outperformed the result of Baseline2 which won the first place VQA Challenge 2016. The idea of combining two paths proves to be effective when using different kinds of image features for abstract images as well. Although many works have been published for VQA, few works have tackled the task with abstract image dataset. We suspect that this is because the abstract scenes dataset is too small to construct a large network architecture, which frequently includes attention mechanism. Due to the small number of training samples, training complex network can decrease the performance. On the other hand, in our model, the architecture is so simple that our model is not influenced by the limitation of samples. Figure 5 shows examples of questions and generated answers in abstract scenes along with the images.

\section{Conclusion}
We implemented DualNet to efficiently and fully account for discriminative information in images and textual features by performing separate operations for input features and building ensemble with varying dimensions. Experiment results demonstrate that DualNet outperforms many previous state-of-the-art results and that it is applicable to both real images and abstract scenes despite their fundamentally different characteristics. In particular, we were able to outperform our own previous state-of-the-art results on abstract scenes category, which recently won the first place at VQA Challenge 2016. Since our method was able to perform well even without attention mechanism, it will be an interesting future work to examine the combination of DualNet and attention mechanism.

\section{ Acknowledgments}
This work was funded by ImPACT Program of Council for Science, Technology and Innovation (Cabinet Office, Government of Japan).

\bibliographystyle{aaai}
\bibliography{bbb}


\end{document}